\documentclass{arxiv_cls}

\title{Utilizing Adversarial Targeted Attacks\\ to Boost Adversarial Robustness}

\author{%
Uriya Pesso \\
School of Electrical Engineering\\
Tel Aviv University\\ 
\texttt{uriyapes@gmail.com}
\AND
Koby Bibas \\
School of Electrical Engineering\\
Tel Aviv University\\
\texttt{kobybibas@gmail.com}
\And
Meir Feder \\
School of Electrical Engineering\\
Tel Aviv University\\ 
\texttt{meir@eng.tau.ac.il}}
\addinstitution{
 School of Electrical Engineering\\
 Tel Aviv University
}

\usepackage{adjustbox}
\usepackage[T1]{fontenc}
\usepackage[utf8]{inputenc}
\usepackage{booktabs}
\usepackage{soul}
\usepackage{pbox}
\usepackage{multirow}

\def\pNMLSingle{
p_{\hat{\theta}(\mathcal{D}_N,x,y)} (y|x)}

\newcommand{\secref}[1]{Section~\ref{#1}} 
\def\secref#1{section~\ref{#1}}
\def\figref#1{figure~\ref{#1}}
\def\Figref#1{Figure~\ref{#1}}

\def\eqref#1{equation~\ref{#1}}
\newcommand\tableref{Table~\ref}

\begin{document}

\maketitle

\begin{abstract}
Adversarial attacks have been shown to be highly effective at degrading the performance of deep neural networks (DNNs).
The most prominent defense is adversarial training, a method for learning a robust model.
Nevertheless, adversarial training does not make DNNs immune to adversarial perturbations.
We propose a novel solution by adopting the recently suggested Predictive Normalized Maximum Likelihood.
Specifically, our defense performs adversarial targeted attacks according to different hypotheses, where each hypothesis assumes a specific label for the test sample.
Then, by comparing the hypothesis probabilities, we predict the label.
Our refinement process corresponds to recent findings of the adversarial subspace properties.
We extensively evaluate our approach on 16 adversarial attack benchmarks using ResNet-50, WideResNet-28, and a 2-layer ConvNet trained with ImageNet, CIFAR10, and MNIST, showing a significant improvement of up to 5.7\%, 3.7\%, and 0.6\% respectively.
% However, even with adversarial training, DNNs remain vulnerable to adversarial perturbed inputs.
% We demonstrate that our proposed approach leads to enhanced robustness for ImageNet, CIFAR10, and MNIST datasets by up to  5.7\%, 3.7\%, and 0.6\% respectively.
\end{abstract}

%-------------------------------------------------------------------------
\section{Introduction}
DNNs have shown state-of-the-art performance on a wide range of problems~\citep{goodfellow2016deep,DBLP:conf/bmvc/KaufmanBBCH19,DBLP:conf/isbi/BibasWCCG21}.
Despite the impressive performance, it has been found that DNNs are susceptible to adversarial attacks \citep{szegedy2013intriguing,biggio2013evasion}. 
These attacks cause the network to underperform by adding specially crafted noise to the input such that the original and modified inputs are almost indistinguishable. 

Different approaches to improve model robustness against adversarial samples have been suggested~\citep{guo2018countering,samangouei2018defensegan,qin2019adversarial, papernot2016distillation, jakubovitz2018improving, zhang2019defense}. Among them, the best performing approach is adversarial training, which augments the training set to include adversarial examples~\citep{goodfellow2014explaining, madry2017towards}.
However, current methods are still unable to achieve a robust model for high-dimensional inputs.

To produce a robust defense, we exploit the \textit{individual setting}~\citep{merhav1998universal}.
In this setting, no assumption is made about a probabilistic connection between the data and labels.
The absence of assumption means that this is the most general framework:
The relationship between the data and labels can be deterministic and may be determined by an adversary that attempts to deceit the model. The generalization error in this setting is referred to as the \textit{regret}.
This regret is the log-loss difference between a learner and the \textit{reference learner}: a learner that knows the true label but is restricted to use a model from a given hypothesis class. 

The pNML learner~\citep{fogel2018universal} was proposed as the min-max solution of the regret, where the minimum is over the model choice and the maximum is for any possible test label value.
The pNML was developed for linear regression~\citep{bibas2019new,DBLP:journals/corr/abs-2102-07181} and evaluated empirically for DNN~\citep{bibas2019deep}.

We propose the \textit{Adversarial pNML} scheme as a new adversarial defense.
Intuitively, the Adversarial pNML procedure assigns a probability for a potential outcome as follows: Assume an arbitrary label for a test sample and perform an adversarial targeted attack toward this label. Take the probability it gives to the assumed label.
Follow this procedure for every label and normalize to get a valid probability assignment.
This procedure can be applied effectively to any adversarial trained model.

To summarize, we make the following contributions.
\begin{enumerate}
    \item We introduce the Adversarial pNML, a novel adversarial defense that enhances robustness.
    This is attained by comparing a set of hypotheses for the test sample. Each hypothesis is generated by a weak targeted attack towards one of the possible labels.  
    \item We analyse the proposed defense and show it is consistent with the recent findings regarding the properties of adversarial subspace. We demonstrate the refinement technique on synthetic data with a simple multilayer preceptron.
    \item 
    We evaluate our approach effectiveness and compare it to state-of-the-art techniques using black-box and white-box attacks, including a defense-aware attack. We show that against white-box untargeted $l_{\infty}$ attack we improve leading methods by $5.7\%$, $3.7\%$ and 0.6\% on ImageNet~\citep{deng2009imagenet}, CIFAR10~\citep{krizhevsky2014cifar} and MNIST~\citep{lecun2010mnist} sets respectively.
\end{enumerate}

Our scheme is simple, requires only one hyper-parameter, and can be easily combined with any adversarial pretrained model to enhance its robustness.
Furthermore, our suggested method is theoretically motivated for the adversarial attack scenario since it relies on the individual setting in which the relation between the data and labels can be determined by an adversary.
Contrary to existing methods that attempt to remove the adversary perturbation~\citep{samangouei2018defensegan,song2018pixeldefend,guo2018countering}, our approach is unique since it does not remove the perturbation but rather use targeted adversarial attack as a defense mechanism.

% -------------- %
% Related Work
% -------------- %
\section{Related work} \label{sec:related_works}
In this section, we mention common adversarial attack and defense methods.

\textbf{Attack methods.}
One of the simplest attacks is Fast Gradient Sign Method (FGSM) \citep{goodfellow2014explaining}. 
Let $w$ be the parameters of a trained model, $x$ be the test data, $y$ its corresponding label, $L$ the loss function of the model, $x_{adv}$ the adversary input, and $\epsilon$ specifies the maximum $l_{\infty}$ distortion such that $||x-x_{\textit{adv}}||_{\infty}\leq \epsilon$. 
First, the signs of the loss function gradients are computed with respect to the image pixels.
Then, after multiplying the signs by $\epsilon$, they are added to the original image to create an adversary untargeted attack
\begin{equation} \label{eq:fgsm_untargeted}
x_{adv} = x + \epsilon \cdot \textit{sign}\nabla_x L(w, x, y_\textit{true}).
\end{equation}
It is also possible to improve classification chance for a certain label $y_\textit{target}$ by performing a targeted attack
\begin{equation} \label{eq:fgsm_targeted}
x_{adv} = x - \epsilon \cdot  \textit{sign}\nabla_x L(w, x,y_\textit{target}).
\end{equation}
A multi-step variant of FGSM was used by \citet{madry2017towards} and is called Projected Gradient Descent (PGD). It is considered to be one of the strongest attacks. 
Denote $\alpha$ as the size of the update, for each iteration an FGSM step is executed
\begin{equation} \label{eq:pgd_untargeted}
x_{\textit{adv}}^{t + 1} = x_{\textit{adv}}^{t} + \alpha \cdot \textit{sign} \nabla_x L(w, x,y_\textit{true}), \quad  0 \leq t \leq T.
\end{equation}
The number of iterations $T$ is predetermined. 
For each sample, the PGD attack is initialized by a random starting point
$x_{\textit{adv}}^{0} = x + u$ where $u \sim U[-\epsilon,\epsilon]$
and the sample $x_{\textit{adv}}^{T}$ with the highest loss is chosen.
% creates multiple different adversarial samples by randomly choosing multiple different starting points $x_{\textit{adv}}^{0} = x + u$ where $u \sim U[-\epsilon,\epsilon]$.
% Lastly, the sample with the highest loss is chosen.
% For each iteration, the maximal allowed distance between the original input and the adversarial input is  $||x - x_{\textit{adv}}^{t}||_{\infty} \leq \epsilon$.
 
% In a black-box scenario, where only the final decision of the model is available to the adversary, decision-based attacks are frequently used. In these attacks, which include Hop-Skip-Jump-Attack (HSJA) \citep{chen2020hopskipjumpattack} among others, the adversary finds a boundary point between the adversarial and non-adversarial region and then move along that boundary to minimize the distance to the original sample \citep{brendel2017decision}.

A different approach is taken by Hop-Skip-Jump-Attack (HSJA) \citep{chen2020hopskipjumpattack}. This is a black-box attack in which the adversary has only a limited number of queries to the model decision. 
The attack is iterative and involves three steps:
estimating the gradient direction, step-size search via geometric progression, and boundary search via a binary search.

\textbf{Defence methods.}
The most prominent defense is adversarial training which augments the training set to include adversarial examples~\citep{goodfellow2014explaining}. Many improvements in adversarial training were suggested. \citet{madry2017towards} showed that training with PGD adversaries offered robustness against a wide range of attacks. \citet{carmon2019unlabeled} suggested using semi-supervised learning with unlabeled data to further improve robustness. \citet{Wong2020Fast} offered a way to train a robust model with a lower computational cost with weak adversaries. 

An alternative to adversarial training is to encourage the model loss surface to become linear such that small changes at the input would not change the output greatly. \citet{qin2019adversarial} demonstrated using a local linear regularizer during training creates a robust DNN model.

% An alternative to adversarial training is to encourage the model loss surface to become linear, so small changes at the input would not change the output greatly. \cite{qin2019adversarial} demonstrated how using a local linear regularizer during training creates a robust DNN model.

% -------------- %
% Preliminaries
% -------------- %
\section{Preliminaries}
\label{sec:preliminaries}
% In supervised machine learning, the goal of the learner is to find an hypothesis which minimize some loss function over the training set $\mathcal{D}_N = \{(x_n,y_n)\}_{n=1}^{N}$ where $x_n \in {\cal X}$ is the $n$-th data sample and $y_n \in {\cal Y}$ is its label. 
% Denote  $\Theta$ as a general index set, the possible hypotheses are a set of conditional probability distributions
% \begin{equation} \label{eq:hypotesis_set}
% P_\Theta = \{ p_\theta(y|x),\;\;\theta\in\Theta\}
% \end{equation}

In supervised machine learning, a training set $\mathcal{D}_N$ consisting of $N$ pairs of examples is given. 
The goal of a learner is to predict the unknown test label $y \in {\cal Y}$ of given new test data $x \in {\cal X}$ by assigning a probability distribution $q(\cdot|x)$ to the unknown label. 
For the problem to be well-posed, we must make further assumptions on the class of possible models or {\em hypothesis set} that is used to find the relation between $x$ and $y$.
Denote  $\Theta$ as a general index set, the possible hypotheses are a set of conditional probability distributions
\begin{equation} \label{eq:hypotesis_set}
P_\Theta = \{ p_\theta(y|x),\;\;\theta\in\Theta\}.
\end{equation}

An additional assumption required to solve the problem is related to how the data and the labels are generated. 
In this work we consider the individual setting \citep{merhav1998universal},
where the data and labels, both in the training and test, are specific individual quantities: We do not assume any probabilistic relationship between them, the labels may even be assigned in an adversarial manner.  
In this framework, the goal of the learner is to compete against a reference learner with the following properties: (i) knows the test label value, (ii) is restricted to use a model from the given hypotheses set $P_\Theta$, and (iii) does not know which of the samples is the test. 
This reference learner then chooses a model that attains the minimum loss over the training set and the test sample
\begin{equation} \label{eq:genie_pnml_org} 
\hat{\theta}(\mathcal{D}_N,x,y)  = \arg\min_{\theta \in \Theta} \left[ -\log p_\theta(y|x) - \sum_{\tiny{(x_n,y_n) \in \mathcal{D}_N}} \log p_\theta(y_n|x_n) \right] .
\end{equation}
The log-loss difference between a learner $q$ and the reference is the regret
\begin{equation} \label{eq:regret}
R(q;\mathcal{D}_N,x,y) = \log \frac{p_{\hat{\theta}(\mathcal{D}_N,x,y)}(y|x)}{q\left(y|x\right)}.
\end{equation} 
The pNML~\citep{fogel2018universal} learner minimizes the regret for the worst case test label
\begin{equation} \label{eq:minmax_prob}
\Gamma = R^*(\mathcal{D}_N,x) = \min_q \max_{y \in \mathcal{Y}} R(q;\mathcal{D}_N,x,y).
\end{equation}
The pNML probability assignment and regret are
\begin{equation} \label{eq:pNML} 
q_{\mbox{\tiny{pNML}}}(y|x)=\frac{\pNMLSingle}{\sum_{y\in {\cal Y}} \pNMLSingle }
, \quad 
\Gamma = \log \sum_{y\in {\cal Y}} \pNMLSingle.
\end{equation}
The pNML regret is associated with the model complexity~\citep{zhang2012model}. This complexity measure formalizes the intuition that a model that fits almost every data pattern very well would be much more complex than a model that provides a relatively good fit to a small set of data.
Thus, the pNML incorporates a trade-off between goodness of fit and model complexity.% as measured by the regret.

% As advocated in \citet{fogelfeder2018}, the chosen universal learner solves:
% \begin{equation} \label{eq:minmax_prob}
% \Gamma = R^*(\mathcal{D}_N,x) = \min_q \max_y R(q;\mathcal{D}_N,x,y).
% \end{equation}

% This min-max optimal solution, termed pNML, is obtained using ''equalizer'' reasoning, following \citep{shtar1987universal}:
% \begin{equation} \label{eq:pNML}
% q_{\mbox{\tiny{pNML}}}(y|x;\mathcal{D}_N)=\frac{\pNMLSingle}{\sum_{y\in {\cal Y}} \pNMLSingle }.
% \end{equation}
% Its corresponding regret, independent of the true $y$, is:
% \begin{equation} \label{eq:pNML_regret}
% \Gamma = \log \sum_{y\in {\cal Y}} \pNMLSingle.
% \end{equation}

% \st{The pNML solution can better handle out-of-distribution test samples since it inherently considers all possible labels when attempting to optimize the worst-case scenario. citet{bibas2019deep} showed that pNML achieves lower loss values.} A more comprehensive analysis on universal learning and pNML can be found in \cite{fogelfeder2018,bibas2019new}. 

\section{Adversarial pNML} \label{sec:adversarial_pNML}
We utilize the pNML learner which is the min-max regret solution of the individual setting. In the individual setting there is no assumption of probabilistic connection between the training and test therefore the result holds for the adversary attack scenario.

% We propose to construct the pNML hypothesis set (\eqref{eq:hypotesis_set}) by adding a refinement stage. This stage uses a pretrained DNN $w$ and modifies $x$ to better fit certain label $y_i$.
% The refinement stage alters the test sample by performing a targeted attack based on the DNN model and arbitrary label $y_i$. Denote $\lambda$ as the refinement strength, the refined sample is:

We propose to construct the pNML hypothesis set (\eqref{eq:hypotesis_set}) with a refinement stage.
Given a pretrained DNN $w$, the refinement stage alters the test sample $x$ by performing a targeted attack toward label $y_i$. Denote $\lambda$ as the refinement strength, the refined sample is
\begin{equation} \label{eq:_x_refine}
x_\textit{refine}(x,y_i) = x - \lambda \cdot \textit{sign}(\nabla_xL(w,x,y_i)).
\end{equation}
This refinement process is repeated for every possible test label. The refined samples are then fed to the pretrained model to compose the hypothesis class
\begin{equation} \label{eq:hypo_class_adv}
P_\Theta = \left\{ p_{w}(\cdot|x_\textit{refine}(x,y_i)), \quad \forall y_i \in \mathcal{Y} \right\}.
\end{equation}
Each member in the hypothesis class produces a probability assignment. 
% We take only the refined label probability for each hypothesis:
In the pNML process we take only the probability it gives to the assumed label
\begin{equation}
p_i = p_{w}(y_i|x_\textit{refine}(x,y_i)).
\end{equation}
We then normalize the probabilities and return the adversarial pNML probability assignment
\begin{equation}
q_\textit{pNML}(y_i) = \frac{p_i}{\sum_{j=1}^{|\mathcal{Y}|} p_j } .
\end{equation}
Since the  refinement is a weak targeted attack, we utilize  a pretrained adversarial trained model to preserve the natural accuracy.
% The corresponding regret is the log normalization factor and is $\Gamma = \log K = \log \sum_{i=1}^{|Y|} p_i$

% Our Adversarial pNML scheme consists of the following steps:
% At first, we train a DNN model $w$ with adversarial training. 
% Then, we produce the hypothesis class -  
% given a test data $x$, we refine it using the trained DNN $w$ and an arbitrary test label $y_i$ (\eqref{eq:_x_refine}).
% We save the label probability we refined with by feeding the refined test data to the trained DNN
% \begin{equation}
% p_i = p_{w}(y_i|x_\textit{refine}(x,y_i)).
% \end{equation}
% This action is repeated for every possible test label value.
% At the end of the process we get a set of predictions, we normalize them and return the Adversarial pNML probability assignment (prediction):
% \begin{equation}
% q_\textit{pNML}(y_i) = \frac{1}{K} p_i, \quad K=\sum_{i=1}^{|Y|} p_i 
% \end{equation}
% The corresponding regret is the log normalization factor and is $\Gamma = \log K = \log \sum_{i=1}^{|Y|} p_i$

%%%%%%%%%%%%%
% Adversarial Subspace Interpretation
%%%%%%%%%%%%%
\subsection{Adversarial subspace interpretation}
\label{subsec:adv_subspace}
We analyze the hypothesis class choice using adversarial subspace properties. 

Let $x_\textit{adv}$ be a strong adversarial example with respect to the label $y_\textit{target}$, i.e.,
the model has a high probability of mistakenly classifying $x_\textit{adv}$ as $y_\textit{target}$.
For the binary classification task there are two members in our suggested hypothesis class:
refinement towards the true label $y_\textit{true}$ and refinement towards the adversary target $y_\textit{target}$. 
There are two mechanisms for strong adversarial examples that cause the refinement towards $y_{true}$ to be stronger than the refinement towards $y_{target}$: convergence to local maxima and refinement overshoot.

\textbf{Convergence to local maxima.} 
\citet{szegedy2013intriguing} stated that adversarial examples represent low-probability pockets in the manifold which are hard to find by randomly sampling around the given sample.
\citet{madry2017towards} showed that FGSM often fails to find an adversarial example while PGD with a small step size succeeds. This implies that for some dimensions the local maxima of the loss is in the interval $[-\epsilon,\epsilon]$. This was also confirmed empirically for CIFAR10 by \citet{Wong2020Fast}.
This means that for some dimensions the local maxima of the loss can be viewed as a ``hole'' in the probability manifold.
For those dimensions, refinement towards $y_{target}$ would not increase the probability of $y_{target}$ hypothesis since $x_{adv}$ already converged to the local maximum. On the other hand, refinement towards $y_{true}$ could cause the refined sample to escape the local maximum hole, thus increasing the probability of $y_{true}$ hypothesis. 

% Following these results, we conclude that for some dimensions the loss local maxima can be viewed as a ``hole'' in the probability manifold.
% For those dimensions, refinement towards $y_{target}$ would not increase the probability of $y_{target}$ hypothesis since $x_{adv}$ already converged to the local maximum. On the other hand, refinement towards $y_{true}$ could cause the refined sample to escape the local maximum hole, thus increasing the probability of $y_{true}$ hypothesis. 

% Overshoot
\textbf{Refinement overshoot.}
\label{refinement_overshoot}
PGD attack is able to converge to strong adversarial points by using multiple iterations with a small step size. This process avoids the main FGSM pitfall: As the perturbation size increases, the gradient direction change \citep{madry2017towards}, causing FGSM to move in the wrong direction and overshoot.
For the same reason, the FGSM refinement towards the $y_{target}$ might fail to create a strong adversarial. 

The refinement towards $y_{true}$ is more probable to succeed since the volume of the non-adversarial subspace is relatively large, thus a crude FGSM refinement is more likely to move in the right direction. To support that claim we note that the adversarial subspace has a low probability and is less stable compared to the true data subspace \citep{tabacof2016exploring}.
In other words, while the true hypothesis escapes the adversarial subspace, the target hypothesis can transform the strong PGD adversarial into weak FGSM adversarial. 
% We show empirically in \secref{sec:experiments} that replacing FGSM refinement with PGD refinement decreases the robustness of our method.

In the case of multi-label classification there is a third kind hypothesis: A refinement towards other label $y \not \in \{y_\textit{true},y_\textit{target}\}$.
This refinement effectively applies a weak targeted attack towards a specific label $y$. 
This hypothesis can be neglected for a strong adversarial input since a weak refinement towards other labels is unlikely to become more probable than refinement towards the target label.

\subsection{Toy example}
\label{subsec:toy_example}
We present an experiment with two-dimensional synthetic data that demonstrate the mechanisms of \secref{subsec:adv_subspace}.

Let $\rho_0 \sim \mathcal{N}(0,0.01I)$ be the distribution with label $0$ and denote $\rho_1 \sim \mathcal{N}(M,0.01I)$ as the distribution of the data the corresponds to label 1. $M$ is a random variable uniformly distributed on a circle of radius 2. 
We train a simple 4 fully connected layer classifier using an adversarial training set generated by a PGD attack with $4$ iterations of size $0.25$ and $\epsilon=0.5$. 
For the adversarial test set, we set $\epsilon$ to $0.95$. The refinement strength $\lambda$ is 0.6.

\begin{figure}[bt]
\begin{tabular}{cc}
\bmvaHangBox{\includegraphics[width=6.0cm]{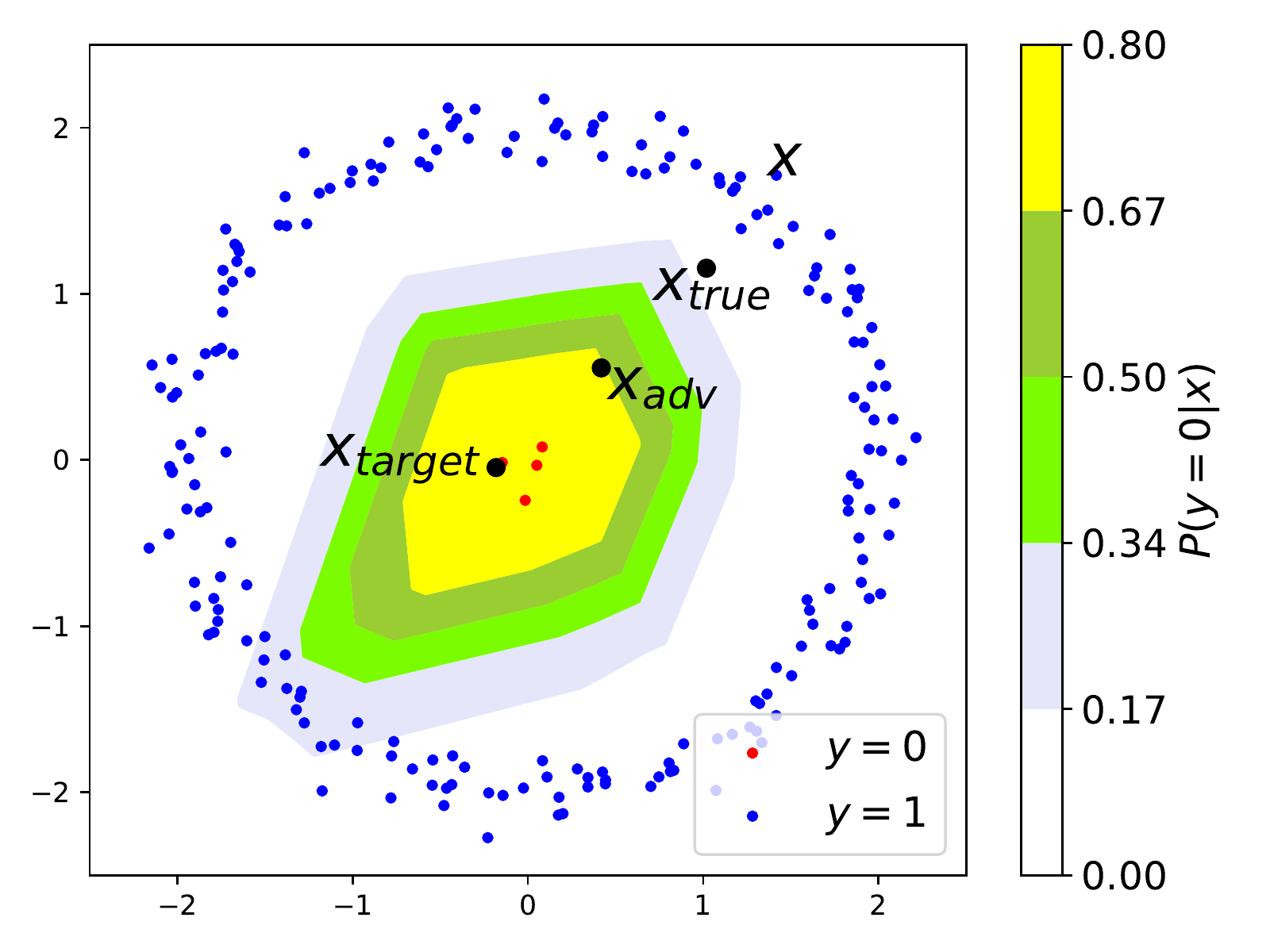}}
&
\bmvaHangBox{\includegraphics[width=6.0cm]{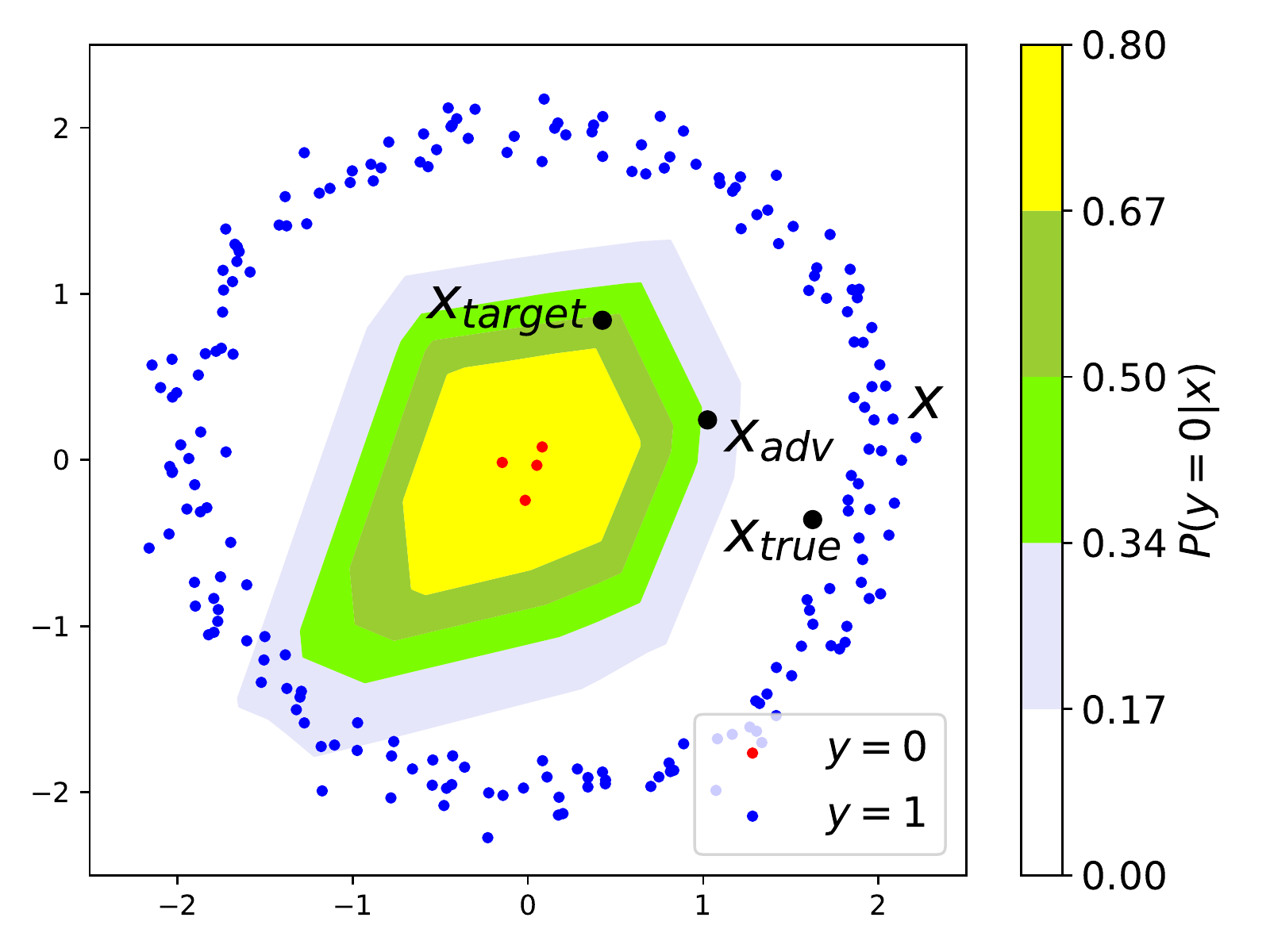}}\\
(a) Convergence to local maxima 
&
(b) Refinement overshoot \\
% \hspace{2.24mm}
\end{tabular}
\caption{
 Label $0$ probability manifold of a 4 fully connected trained model.
 We use synthetic data: label 0 samples were created from $\mathcal{N}(0,0.01I)$ and label 1 data were generated from $\mathcal{N}(M,0.01I)$ where $M$ is a random variable uniformly distributed on a circle of radius 2. This figures demonstrate the mechanisms that cause the refinement towards $y_{true}=1$ to be stronger than the refinement towards $y_{target}=0$.}
 \label{fig:toy_ds_demo}
\end{figure}

\Figref{fig:toy_ds_demo} shows the refinement process overlaid on the trained model label $0$ probability manifold. $x$ is the original sample with label $1$, $x_{adv}$ is the test adversarial sample, $x_{true}= x_{\textit{refined}}(x_\textit{adv},y=1)$ is the sample generated by refinement towards label $1$, and $x_{target}=x_{\textit{refined}}(x_\textit{adv},y=0)$ is generated by refinement towards label $0$.

\Figref{fig:toy_ds_demo}\textcolor{red}{a} demonstrates the convergence to local maxima mechanism.
$x_{adv}$ converged to the maximum probability, therefore refinement towards the target label does not increase the probability while refinement towards the true label does. As a result, the true hypothesis probability is greater and the true label is predicted.
\Figref{fig:toy_ds_demo}\textcolor{red}{b} presents the refinement overshoot mechanism. The target hypothesis is refined in the wrong direction while the true hypothesis is refined in the correct direction.
This makes the Adversarial pNML prediction to be more robust to adversarial attacks.

%------------------------ %
% Experiments
%------------------------ %
\section{Experiments}  \label{sec:experiments}
In this section, we present experiments that test our proposed Adversarial pNML scheme as a defense for adversarial attack. 
We evaluate the natural performance (performance on images without perturbation) and adversarial performance on MNIST~\citep{lecun2010mnist}, CIFAR10~\citep{krizhevsky2014cifar} and ImageNet~\citep{deng2009imagenet} datasets. We compare our scheme to recent leading methods.

% ------------------- %
% Adaptive Adversary
% ------------------- %
\subsection{Adaptive attack and gradient masking} 
\label{subsec:adaptive_adversary}

A main part of the defense evaluation is creating and testing against adaptive adversaries that are aware of the defense mechanism~\citep{carlini2019evaluating}. This is specifically important when the defense cause gradient masking~\citep{papernot2017practical}, in which gradients are manipulated, thus prevent a gradient-based attack from succeeding. 
Defense aware adversaries can overcome this problem by using a black-box attack or by approximating the true gradients~\citep{athalye2018obfuscated}. 

We design a defense-aware adversary for our scheme: We create an end-to-end model that calculates all possible hypotheses in the same computational graph.
We note that the end-to-end model causes gradient masking since the refinement $sign(\cdot)$ function sets some of the gradients to zero during the backpropagation phase. 
We, therefore, attack the end-to-end model with the black-box HSJA method and PGD with Backward Pass Differentiable Approximation (BPDA) technique~\citep{athalye2018obfuscated}, denoted as an adaptive attack.

In BPDA we replace the non-differentiable part with some differentiable approximation on the backward pass. Assuming the refinement is small, one solution is simply to approximate the refinement stage by the identity operator $x_\textit{refine}\approx x$, which leads to $\frac{\partial x_\textit{refine}}{\partial x} \approx 1$. 
Further discussion on the adaptive attack can be found in the appendix.

% -------------------- %
%    Table Results
% -------------------- %
\begin{table}[bt]
\begin{center}  
\small
\begin{tabular}{c|c|c|cccc|c}
\toprule
Dataset & Method & Natural & FGSM  & PGD & Adaptive & HSJA  & Best attack \\
\toprule
\multirow{3}{*}{\shortstack[l]{MNIST \\ $\epsilon=0.3$}}
& Standard & 99.3\%  & \ 0.6\%  & 0.0\% & - & - & 0.0\% \\
& \citet{madry2017towards} & 97.8\% & 95.4\%  & 91.2\% & 89.8\% & 93.1\% & 89.8\% \\
& Ours & 97.8\%  & 95.4\%  & 93.7\% & 90.4\% & 94.6\% & \textbf{90.4\%} \\
\midrule 
\multirow{5}{*}{\shortstack[l]{CIFAR10 \\ $\epsilon=0.031$}}
&  Standard & 93.6\%  & 6.1\% &  0.0\% & - & - & 0.0\% \\
% &  WRN & 83.7\%  & 48.2\%  &  37.4\% & - & - & 37.4\% \\
    %   & WRN + ours  & 84.3\%  &  48.9\% & 45.7\% & - & - &  45.7\% \\ \
&  \citet{madry2017towards} & 87.3\%  & 56.1\%  &  45.8\% & - & - & 45.8\% \\
&  \citet{qin2019adversarial} & 86.8\%  & - &  54.2\% & - & - & 54.2\% \\
&  \citet{carmon2019unlabeled} & 89.7\%  & 69.9\%  &  62.7\% & - & 78.8\%  & 62.7\% \\
& Ours & 88.1\%  & 69.5\%  & 67.2\% & 66.4\% & 84.8\% & \textbf{66.4\%} \\
\midrule
\multirow{3}{*}{\shortstack[l]{ImageNet \\ $\epsilon=8/255$}}
&  Standard & 83.5\% & 7.0\% & 0.0\% & - & - & 0.0\% \\
& \citet{Wong2020Fast} & 69.1\%  & 27.0\%  &  16.0\% & -  & 68.0\% & 16.0\% \\
& Ours & 69.3\% & 28.0\% & 20.0\% & 19.0\% & 68.0\% & \textbf{19.0\%} \\
\midrule
\multirow{2}{*}{\shortstack[l]{ImageNet \\ $\epsilon=4/255$}}
& \citet{Wong2020Fast} & 69.1\%  & 44.3\%  &  42.9\% & -  & - & 42.9\% \\
&  Ours & 69.3\% & 49.0\% & 48.6\% & - & - & \textbf{48.6\%} \\
\bottomrule
\end{tabular}
\end{center}
\caption{A comparison of different defense accuracy against various adversarial attacks.}
\label{table:imagenet_results}
\label{table:cifar_results}
\label{table:mnist_results}
\end{table}

\subsection{Experimental results} \label{sec:experiment_results}

% We use a network that consists of two convolutional layers with 32 and 64 filters respectively, each followed by $2\times2$ max-pooling, and a fully connected layer of size 1024. 
% We trained the model for 106 epochs with adversarial trainset that was produced by PGD based attack on the natural training set with 40 steps of size 0.01 with a maximal $\epsilon$ value of 0.3. We used SGD with a learning rate of $0.01$, momentum value 0.9 and weight decay of 0.0001. For the last 6 epochs, we used adversarial training with the adaptive attack instead of PGD.

% We compared our scheme to a base model, i.e., the model without our scheme. To perform a fair comparison, we evaluated the base model against a PGD attack with the same settings. 

% MNIST Dataset
\textbf{MNIST.} \label{subsubsec:mnist_experiments}
We follow the model architecture as described in \citet{madry2017towards}. 
We use a model that consists of two convolutional layers with 32 and 64 filters respectively, each followed by $2\times2$ max-pooling, and a fully connected layer of size 1024 (training details can be found in the appendix).
We set the Adversarial pNML refinement strength to $\lambda=0.1$.

For evaluation, we set the attack strength to $\epsilon=0.3$ for all attacks. The PGD attack was configured with 50 steps of size 0.01 and 20 restarts. For the adaptive attack, we used 300 steps of size 0.01 and 20 restarts. 
For HSJA, we set the number of model queries to $26K$ per sample, which was shown to be enough queries for convergence \citep{chen2020hopskipjumpattack}.

In \tableref{table:mnist_results} we report the accuracy of our scheme in comparison to the adversarial trained model without our scheme (\citet{madry2017towards}). 
We observe that Adversarial pNML improves the robustness by 0.6\% without degrading the accuracy of images with no adversarial perturbation (natural accuracy). The adaptive attack is the best attack against our defence which indicates that this kind of attack is efficient. 
In addition, our scheme improves the accuracy for black-box attack by 1.5\% as seen in \tableref{table:mnist_results} HSJA column.

% For CIFAR10 dataset, we evaluated our scheme using two models:
% (i) a pre-trained Wide-ResNet 28-10 architecture \citep{zagoruyko2016wide}, that was trained with both labeled and unlabeled data \citet{carmon2019unlabeled}; (ii) a similar model, denoted as WRN, trained with PGD adversarial samples (training parameters can be found in the Appendix).
% We set the Adversarial pNML refinement strength to $\lambda=0.03$ for both models. 
% For evaluation, we set $\epsilon=0.031/1$ for all attacks. PGD and adaptive attack were configured with 20 steps of size 0.007 and without random starts. The adaptive attack parameters were determined following a hyperparameters search (we report the results in the Appendix). For HSJA, we set the maximal number of model queries to $26K$ per sample and evaluate the accuracy for the first 2000 samples (out of 10000). 

% We denote $Ours$ as our scheme with \citet{carmon2019unlabeled} base model.

% CIFAR10 Dataset
\textbf{CIFAR10.}
We build our scheme upon a pre-trained WideResNet 28-10 architecture \citep{zagoruyko2016wide} trained by \citet{carmon2019unlabeled} with both labeled and unlabeled data.
We set the Adversarial pNML refinement strength to $\lambda=0.03$. 
For evaluation, we set $\epsilon=0.031$ for all attacks. PGD and adaptive attack were configured with 200 steps of size 0.007 and 5 restarts. 
% The adaptive attack parameters were determined following a hyperparameters search (we report the results in the Appendix). 
For HSJA attack, we set the maximal number of model queries to $26K$ per sample and evaluate the accuracy for $2K$ samples. 

In \tableref{table:cifar_results} we report the accuracy of our scheme in comparison to other state-of-the-art algorithms. 
We observe that our method achieves state-of-the-art performance, enhancing the robustness by 3.7\% with the best natural accuracy when compared to other defenses.
The Adversarial pNML improves the accuracy against black-box attacks by 6.0\%. which shows that the robustness boost of our method is not due to the masked gradients.

For the FGSM attack, \citet{carmon2019unlabeled} outperforms our scheme by 0.4\%. This result, together with the improvement our method achieves against PGD attack demonstrates the convergence to local maxima mechanism as described in \secref{subsec:adv_subspace}. 
% Our scheme works best when the adversarial sample converged to a loss local maxima.  

% We incorporate our suggested Adversarial pNML method with PGD based adversarial trained model as suggested by \citet{madry2017towards} and named it \emph{WRN + ours}.
% In \tableref{table:cifar_results}, one can see that our scheme enhances the robustness of the WRN model by 8\% with no loss in natural accuracy. 

\begin{figure}[bt]
\begin{tabular}{cc}
\bmvaHangBox{\includegraphics[width=6.0cm]{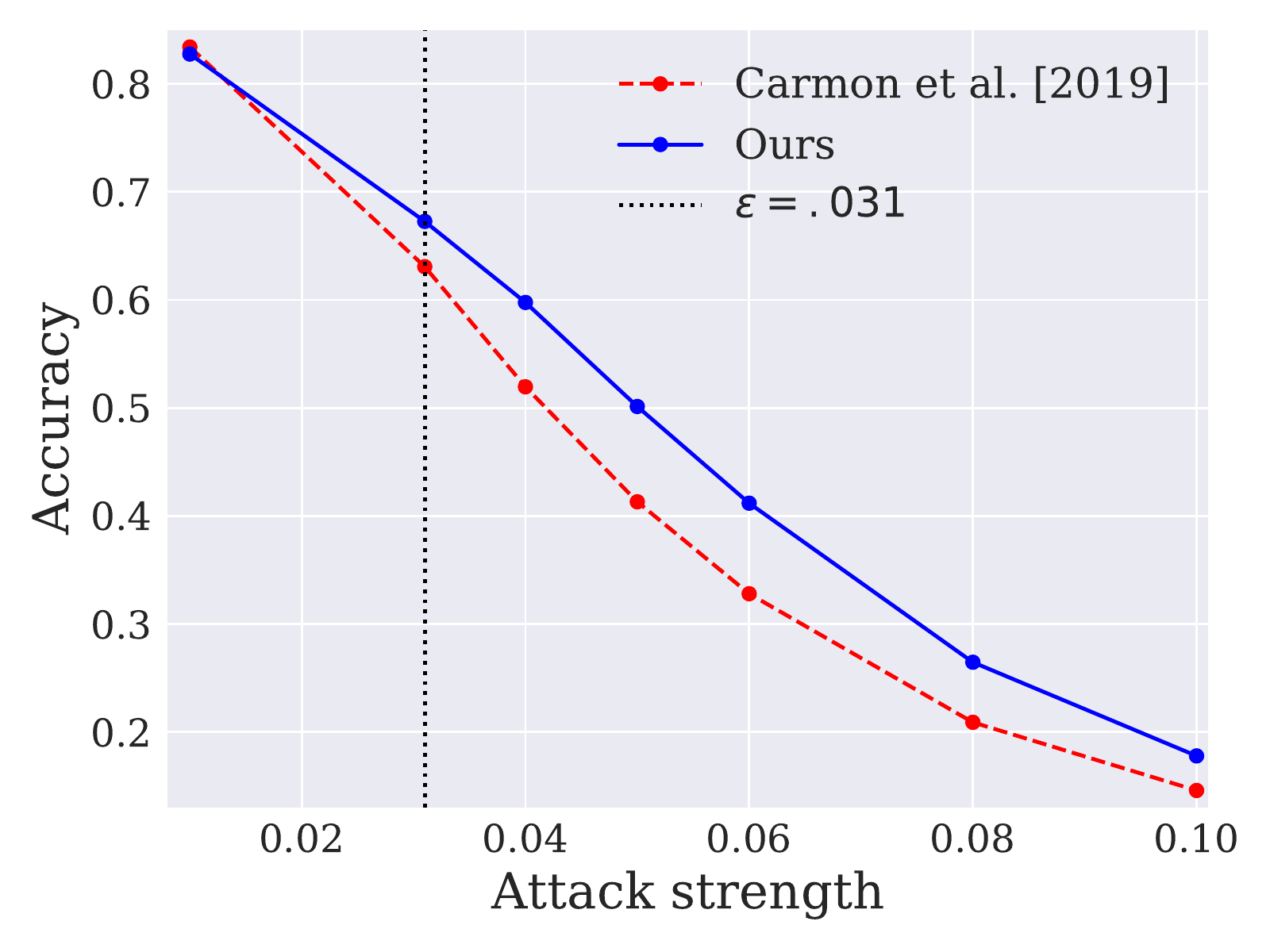}}&
\bmvaHangBox{\includegraphics[width=6.0cm]{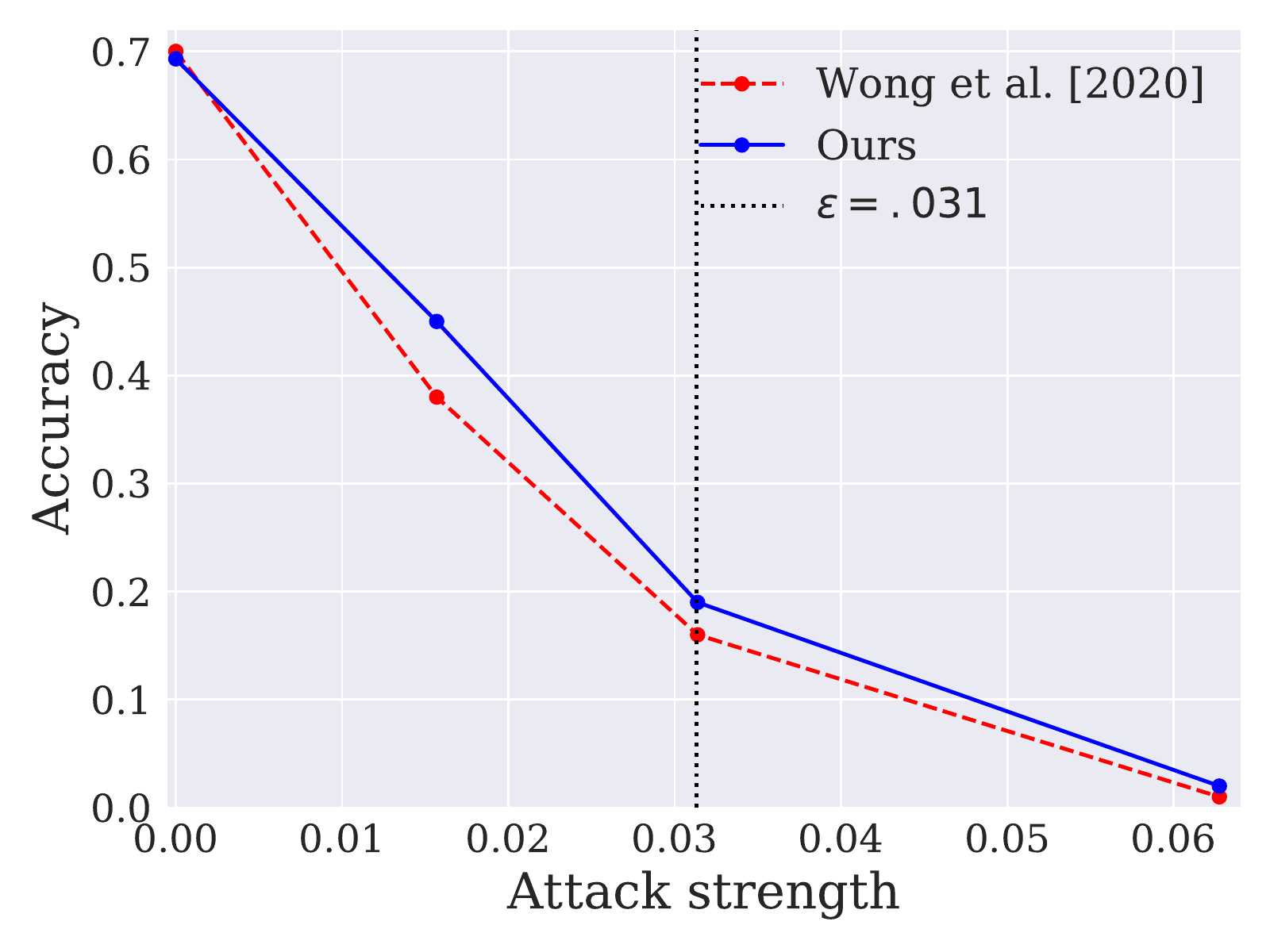}}\\
(a) CIFAR10 &(b) ImageNet
\end{tabular}
\caption{Robustness for different attack strengths ($\epsilon$). We compare the robustness between the adversarially trained model and the same model with our scheme. The comparison is for the best attack of each model. The dashed line marks the $\epsilon$ that was used in \tableref{table:cifar_results}.}
\label{fig:acc_vs_attack_strength}
\label{fig:subset_imagenet_pgd_acc}
\end{figure}

In \figref{fig:acc_vs_attack_strength}\textcolor{red}{a} we show the robustness of our scheme against PGD attack for various of attack strengths. The results show that our approach is more robust for all $\epsilon$ values greater than 0.01. 
Specifically, the maximal improvement is 8.8\% for $\epsilon=0.05$.
For $\epsilon=0.01$ our scheme is less robust by 0.6\%. To explain these results, recall that the refinement strength is 0.03. 
When $\epsilon<<\lambda$, one of the refinement hypotheses could generate adversarial examples stronger than the examples generated by the adversarial attack.

%    ImageNet Dataset
\textbf{ImageNet.}
We utilize a pre-trained ResNet50 trained by \citet{Wong2020Fast} using fast adversarial training with $\epsilon=4/255$. 
We set the Adversarial pNML refinement strength to $\lambda=3/255$. 
We used a subset of the evaluation set containing 100 labels.
PGD and adaptive attack were configured with 50 steps of size $1/255$ and with 10 random restarts. For HSJA, we set the number of model queries to $12K$ per sample. 

In \tableref{table:imagenet_results} we report the accuracy of our scheme in comparison to \citet{Wong2020Fast} for $\epsilon$ values of $8/255$ and $4/255$.
For $\epsilon=4/255$, we evaluated PGD and FGSM attacks using $5K$ samples and for $\epsilon=8/255$ we used 100 samples (1 sample per label).
We observe that robustness is improved by 3\% and 5.7\% for $\epsilon$ values of $8/255$ and $4/255$ respectively. The accuracy on natural images is improved by 0.2\%. 
The HSJA attack seems to fail in finding adversarial examples, and that for $\epsilon=8/255$ the adaptive attack is the best. 

In \figref{fig:subset_imagenet_pgd_acc}\textcolor{red}{b} we explore the robustness of our scheme against adaptive attack. The results show that our scheme is more robust for all $\epsilon\geq0.016$, specifically the maximal improvement is 6.4\% for $\epsilon=0.024$.

\begin{figure}[tb]
\begin{minipage}{0.55\textwidth}
% \flushleft
    \centering
    \bmvaHangBox{\includegraphics[width=\textwidth]{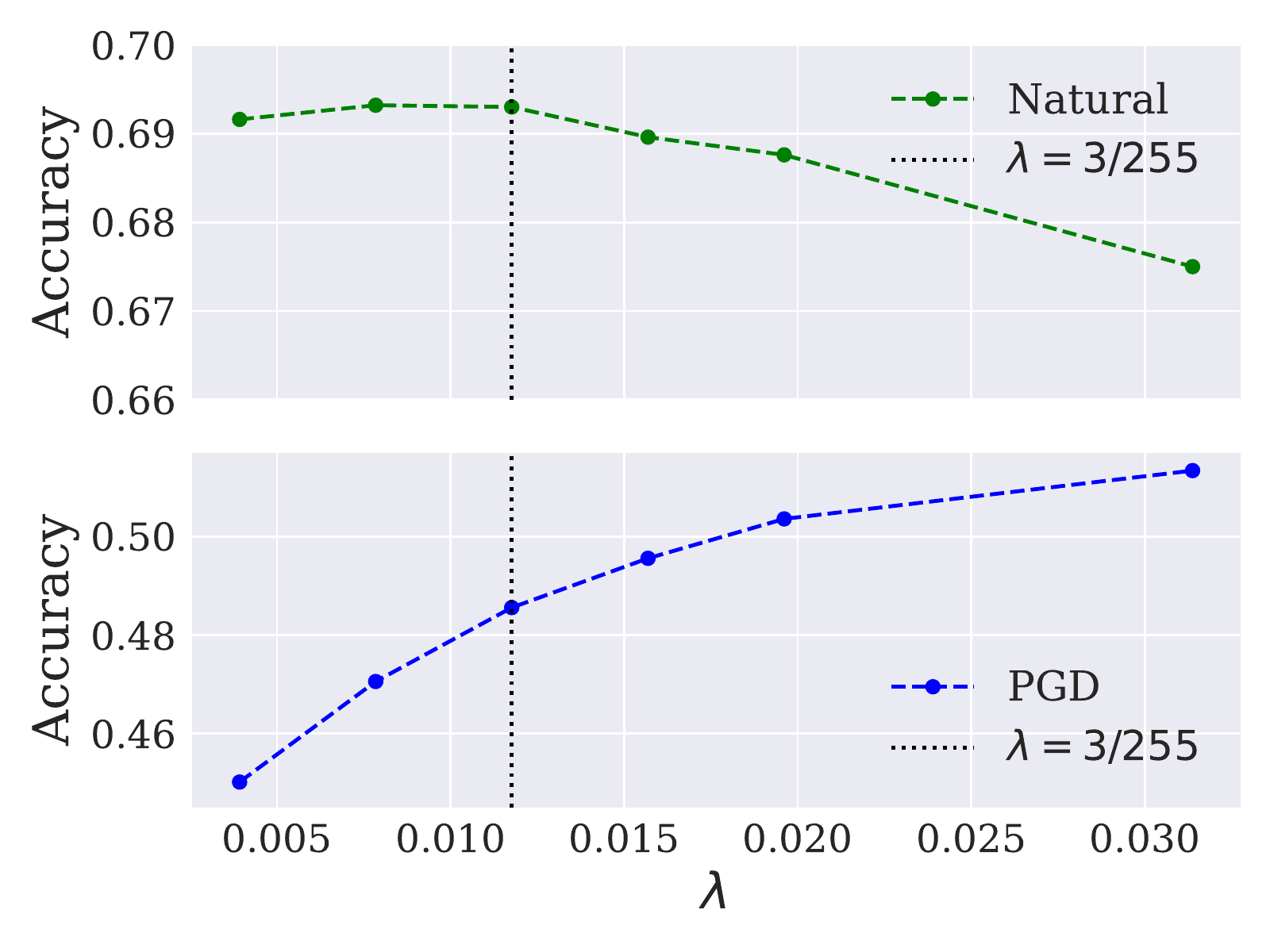}}
    \caption{ImageNet accuracy for different $\lambda$.} 
    \label{fig:imagenet_acc_vs_lambda}
\end{minipage}%
\hfill
\begin{minipage}{0.44\textwidth}
    \small
    \centering
    \begin{tabular}{c|c|c}
        \toprule
        Accuracy & Iterations & Step size \\
        \toprule
        67.26\% & 1 & 0.03 \\
        66.89\% & 2 & 0.015 \\
        66.86\% & 4 & 0.01 \\
        \bottomrule
    \end{tabular}
    \vspace{0.25cm}
    \caption{CIFAR10 accuracy for PGD attack for various refinement parameters. Increasing the iteration number and decreasing the step size makes the refinement process more precise.}
    \label{table:refinement_overshoot}
\end{minipage}
\end{figure}

% \begin{figure}[bt]
% \begin{tabular}{cc}
% \bmvaHangBox{
% \includegraphics[width=7.0cm]{figures/fig_imagenet_acc_vs_lambda.pdf}}
% &
% \begin{small}
% \end{small}
% \small
% \bmvaHangBox{\begin{adjustbox}{width=5cm}                
% \begin{tabular}{c|c|c}
%     \toprule
%     Accuracy & Iterations & Step size \\
%     \toprule
%     67.26\% & 1 & 0.03 \\
%     66.89\% & 2 & 0.015 \\
%     66.86\% & 4 & 0.01 \\
%     \bottomrule
%     \end{tabular}
%     \end{adjustbox}
%     % \\\textbf{Table 2:} CIFAR10 accuracy against\\PGD attack for different refinement\\iterations and refinement step sizes.\\
%     % \\\textbf{Figure 3:} ImageNet accuracy Vs.\\refinement strength $\lambda$
%     }\\
% % (a) ImageNet accuracy Vs. $\lambda$ &(b) CIFAR10 accuracy using different refinement parameters \\
% % \hspace{2.24mm}
% \pbox{7cm}{\textbf{Figure 3:} ImageNet accuracy Vs. 
% \\ refinement strength $\lambda$} & \pbox{5cm}{\textbf{Table 2:} CIFAR10 PGD accuracy \\ for different refinement parameters}
% \end{tabular}
% \label{fig:imagenet_acc_vs_lambda}
% \label{table:refinement_overshoot}
% \end{figure}

\subsection{Ablation study} \label{sec:ablation_study}
The choice of the refinement strength represents a trade-off between the robustness against adversarial attacks and the accuracy of natural samples. 
This trade-off is explored in \figref{fig:imagenet_acc_vs_lambda}.
As the refinement strength increases so is the robustness to PGD attack at the price of a small accuracy loss for natural samples. 
A good choice of the refinement strength $\lambda$ would be in the interval $[0.5\epsilon,\epsilon]$. This gives a good balance between natural and adversarial accuracy. 
In our experiments, we used a small validation set to find a good $\lambda$ value.

% Denote $\epsilon$ as the strength used during the adversarial training,
% from multiple experiments conducted on multiple models, we conclude that usually, a refinement strength in the interval $[0.5\epsilon,\epsilon]$ would give a good balance between natural and adversarial accuracy. 

In \tableref{table:refinement_overshoot} we explore the overshoot mechanism (\secref{subsec:adv_subspace}). We adjust the refinement to become more precise by replacing FGSM refinement with a PGD refinement which uses more iterations and smaller step size. We test our method on CIFAR10 against PGD attack with the same settings described in \secref{sec:experiment_results}. The results show that as the refinement becomes more precise, the robustness against PGD attack decreases. This demonstrates that the overshoot mechanism improves robustness since PGD refinement, which is less prone to overshoot, has lower robustness. This supports the claim of the instability of the adversarial subspace \citep{tabacof2016exploring}, which explains why FGSM refinement towards $y_{true}$ is more likely to succeed compared to refinement towards the $y_{target}$.

\subsection{Run-time analysis}
Let $H$ be the number of hypotheses, i.e., the number of possible test labels.
For each sample, our method performs a forward-pass (FP) followed by $H$ backward-passes (BP) to generate the refined samples. 
Then additional $H$ FP are made to calculate the prediction. It is possible to calculate these values simultaneously by batching, which reduces the complexity to $2\cdot FP + BP$. To decrease the batch size, we can reduce the number of hypotheses with a minor degradation to robustness by only calculating the hypotheses of the most probable labels. More information is available in the appendix.

% This provide further support to the claim that non-adversarial subspace is larger than adversarial subspace, which explain why FGSM refinement towards $y_{true}$ is more likely to succeed compared to refinement towards the $y_{target}$.

% \subsection{Runtime Speed}
% Contrary to adversarial training, our scheme does not require any resources during the training phase. On the other hand, during inference, the refinement requires additional backward pass and forward pass. We calculate all the hypothesis simultaneously using batching but acknowledge it requires more memory.

\section{Conclusion}
We presented the Adversarial pNML scheme for defending DNNs from adversarial attacks. The theory behind this scheme comes from the individual setting where the relation between the data and labels can be determined by an adversary. 
Our method is conceptually simple, requires only one hyper-parameter, and flexible since it allows a trade-off between robustness and natural accuracy. Furthermore, any pretrained model can be easily combined with our scheme to enhance its robustness.
We analysed the mechanisms that enable our method to boost the robustness using properties of the adversarial subspace.
We showed empirically that our method enhances the robustness against adversarial attacks for ImageNet, CIFAR10, and MNIST datasets by $5.7\%$, $3.7\%$, and $0.6\%$ respectively. 

This work suggests several potential directions for future work: The pNML regret, which is the log-loss distance from the reference learner, can form an adversarial attack detector.
In addition, we would like to explore other hypothesis classes as the entire model parameter class where the model weights are changed according to the different hypotheses.

\bibliography{main_bib}
\appendix
\newpage
\section{Training parameters}
We now detail the training parameters and architecture used to train the different models.
\paragraph{MNIST.} For both the standard model and  \citet{madry2017towards} model we used a network that consists of two convolutional layers with 32 and 64 filters respectively, each followed by $2\times2$ max-pooling, and a fully connected layer of size 1024. 

We trained the standard model for 100 epochs with natural training set. We used SGD with a learning rate of $0.01$, a momentum value of 0.9, a weight decay of 0.0001, and a batch size of 50. 

We trained the \citet{madry2017towards} model for 106 epochs with adversarial trainset that was produced by PGD based attack on the natural training set with 40 steps of size 0.01 with a maximal $\epsilon$ value of 0.3. We used SGD with a learning rate of $0.01$, momentum value 0.9 and weight decay of 0.0001. For the last 6 epochs, we used adversarial training with the adaptive attack instead of PGD. We set the Adversarial pNML refinement strength to $\lambda=0.1$.

\paragraph{CIFAR10.} We used wide-ResNet 28-10 architecture \citep{zagoruyko2016wide} for the standard model.
We trained the standard  over 204 epochs using SGD optimizer with a batch size of 128 and a learning rate of 0.001,
reducing it to 0.0001 and 0.00001 after 100 and 150 epochs respectively. We also used a momentum
value of 0.9 and a weight decay of 0.0002. 

\paragraph{ImageNet.} For the standard model we used a pre-trained ResNet50 \citep{he2016deep}. Similarly to the other models, we adjust the standard model to only output the first 100 logits.

\section{Adaptive attack}
\begin{figure}
\centering
\includegraphics[width=1.0\linewidth]{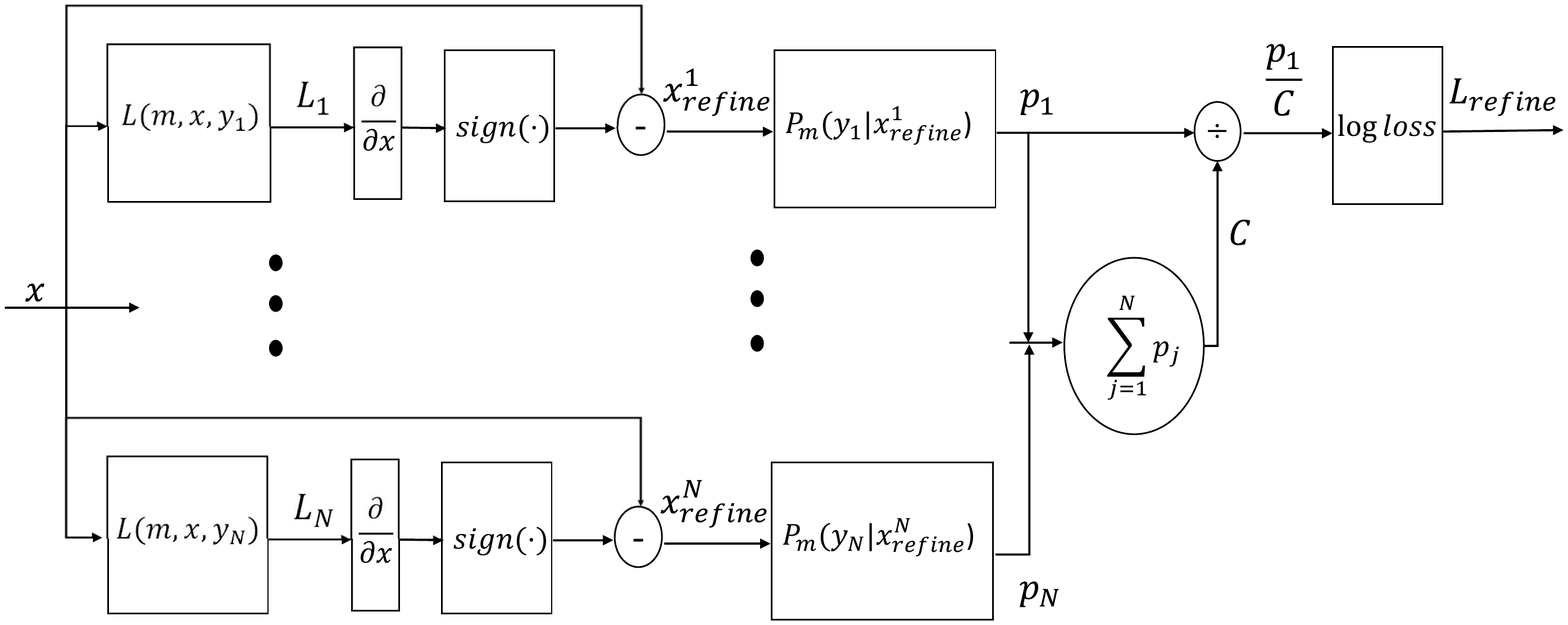}
\caption{End-to-end model illustration presenting label 1 hypothesis log-loss.}
\label{fig:adv_pnml_scheme}
\end{figure}

In this section we discuss alternative approximations for the adaptive attack. \Figref{fig:adv_pnml_scheme} presents the end-to-end model. We denote $(x, y_1)$ as an input that belong to label $y_1$, $w_0$ is the model parameters, and $L(w_0,x,y_i)$ is the model loss w.r.t a specific label $y_i$ where $i \in [1,N]$. $x_{refine}^i$ is the refinement result for the $i$-th hypothesis and $p_i/C$ is the probability of the corresponding hypothesis. $L_{refine}$ is the loss for the first hypothesis. The adaptive adversary manipulate the input by taking steps in the direction of the gradients $\frac{\partial L_{refine}}{\partial x}$.

Recall that our adaptive attack approximates the refinement stage with a unity operator on the backward pass which leads to $\frac{\partial x_\textit{refine}}{\partial x} \approx 1$ (see \secref{subsec:adaptive_adversary}). This approach, in effect, disregard anything that comes before the $sign(\cdot)$ operator during backpropagation. 
An alternative approach is to use some kind of a differentiable function to approximate the $sign(\cdot)$ operator and backpropagate through the entire computational graph. The first obstacle is to find a differentiable function that approximates the $sign(\cdot)$ operator well. The first option that comes to mind is to use a $tanh(\cdot)$ function, but since the input values are distributed across a wide range, the $tanh(\cdot)$ causes a vanishing gradients effect, which misses the goal of this approximation.
% makes this approximation useless for our goal

Another approach is to disregard the $sign(\cdot)$ operator on the backward pass, i.e., backpropagate the gradients without changing them. We examine this case:
% \begin{equation}
%     \frac{\partial L_{refine}}{\partial x} = \frac{\partial log(\frac{p_1}{\sum_{i=1}^{N} p_i})}{\partial x} = \frac{\partial log(p_1)}{\partial x} - \frac{\partial log({\sum_{i=1}^{N} p_i})}{\partial x}
% \end{equation}
% Using the chain rule:
\begin{equation}
    \frac{\partial L_{refine}}{\partial x} = \sum_{i=1}^{N} \frac{\partial log(\frac{p_1}{\sum_{j=1}^{N} p_j})}{\partial p_i} \frac{\partial p_i}{\partial x} = 
    \frac{1}{p_1}\frac{\partial p_1}{\partial x} - \frac{1}{\sum_{j=1}^{N} p_j}\sum_{i=1}^{N}\frac{\partial p_i}{\partial x},
\end{equation}
\begin{equation}
\frac{\partial p_i}{\partial x} = 
\frac{\partial p_{w_o}(y_i|x^i_{refine})}{\partial x} = 
\frac{\partial p_{w_o}(y_i|x^i_{refine})}{\partial x^i_{refine}}\frac{\partial x^i_{refine}}{\partial x},
\end{equation}

\begin{equation} \label{eq:hessian}
\frac{\partial x^i_{refine}}{\partial x} = 
\frac{\partial (x - \frac{\partial L_i}{\partial x})}{\partial x} =
1 - \frac{1}{\partial x}\left(\frac{\partial L_i}{\partial x}\right).
\end{equation}

Note that \eqref{eq:hessian} is dependent on the Hessian matrix of the loss $L_i$ w.r.t $x$. Computing this value is computationally hard for DNN's and it is usually outside the scope of adversarial robustness tests - only first-order adversaries are considered \citep{madry2017towards}. This emphasizes that the only viable, gradient-based, adaptive attack is the one used in our paper 

% Where $\boldsymbol{p} = \left[p_1,p_2,\ldots,p_n \right]$
% \begin{enumerate}
% \item
% $f(\cdot) = 0.5 + 0.5tanh(k\cdot)$ 
% \item
% $f(\cdot) = k\cdot $
% \end{enumerate}
%  that multiply the gradients. 

% As explained in Section \ref{subsec:adaptive_adversary}, the refinement result for each hypothesis input is refined using equation \ref{eq:_x_refine}

\begin{figure}
\centering
\includegraphics[width=0.7\linewidth]{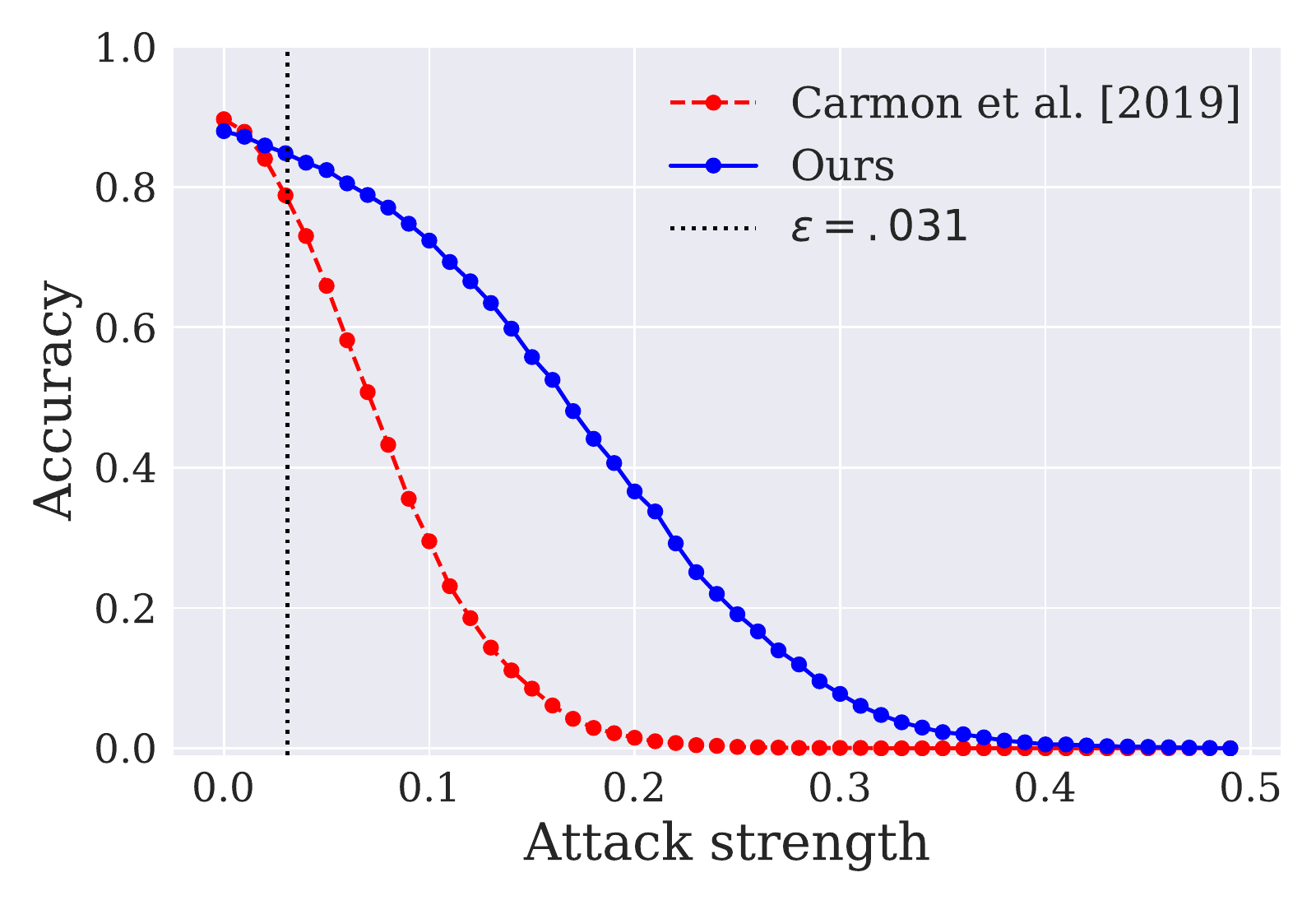}
    \caption{CIFAR10 accuracy against HSJA attack. We compare the robustness between the base model and the same model with our scheme. The dashed line mark the $\epsilon$ that is used in \tableref{table:cifar_results}.}
    \label{fig:cifar_hsja}
\end{figure}

\section{Additional CIFAR10 results}
We now provide additional results that support the claim that our scheme does indeed enhance robustness.

\paragraph{HSJA for different $\epsilon$ values.} In figure \ref{fig:cifar_hsja} we demonstrate that our scheme is more robust against black-box HSJA for various $\epsilon$ values. Specifically, the maximal improvement is 49.1\% for $\epsilon=0.13$. We note that in comparison to the white-box attack, HSJA is much less efficient against our scheme for large $\epsilon$ values. Nevertheless, the improvement of our scheme against black-box attack supports the claim that its robustness enhancement is not only the result of masked gradients.

\section{Run-time analysis}
\begin{figure}
\centering
\includegraphics[width=0.8\linewidth]{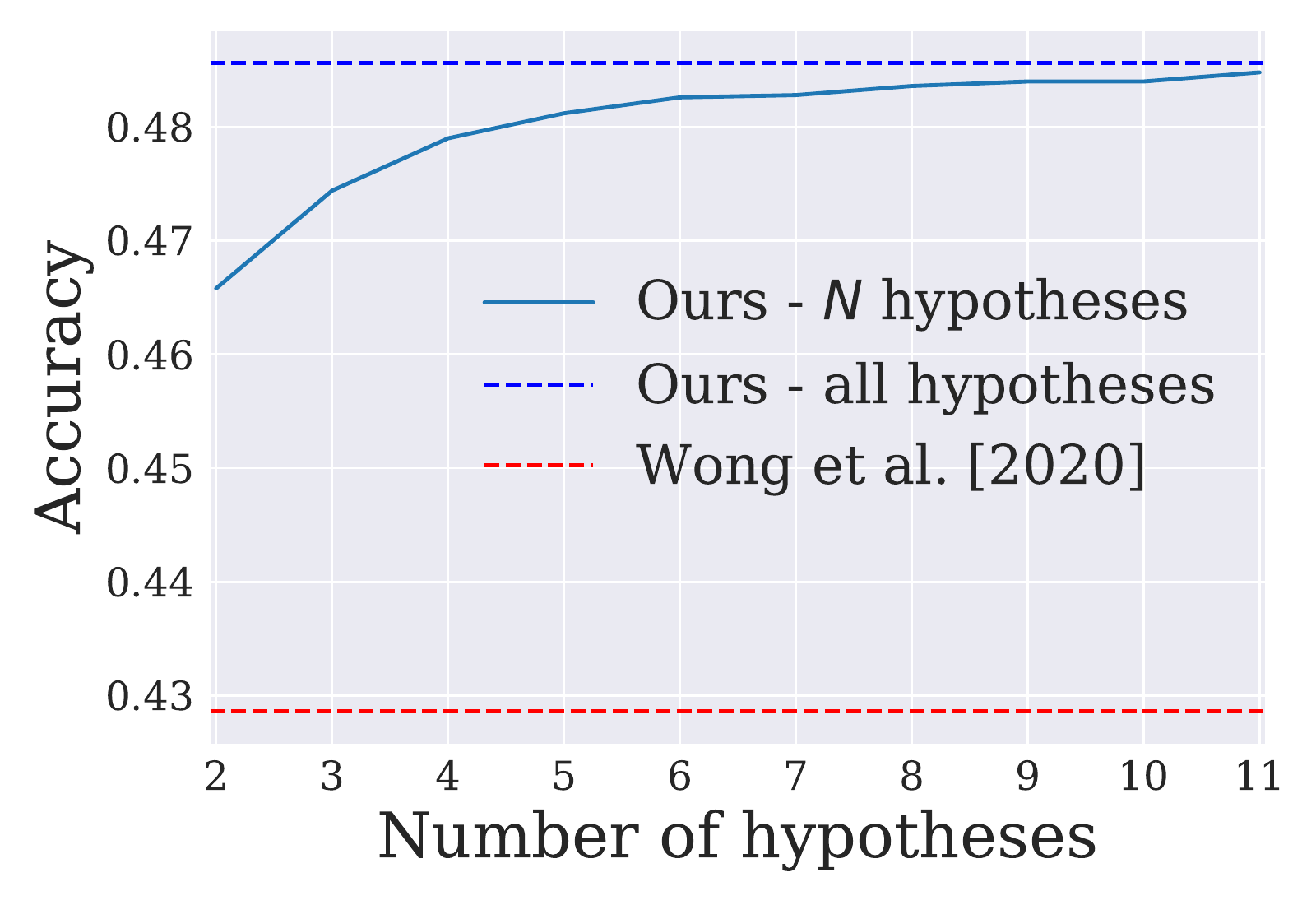}
    \caption{ImageNet accuracy against PGD attack Vs. the Number of hypotheses.}
    \label{fig:imagenet_acc_vs_hypotheses}
\end{figure}

Let $H$ be the number of hypotheses, i.e., the number of possible test labels.
For each sample, our method performs a forward-pass (FP) followed by $H$ backward-passes (BP) to generate the refined samples. 
Then additional $H$ FP are made to calculate the prediction. It is possible to calculate these values simultaneously by batching, which reduces the complexity to $2\cdot FP + BP$. To decrease the batch size, we can reduce the number of hypotheses with a minor degradation to robustness by only calculating the hypotheses of the most probable labels (which we know after the first $FP$), demonstrated in \Figref{fig:imagenet_acc_vs_hypotheses}.

\end{document}